\documentclass{article}

\usepackage[nonatbib]{nips_2017}

\usepackage[utf8]{inputenc} 
\usepackage[T1]{fontenc}    
\usepackage{hyperref}       
\usepackage{url}            
\usepackage{booktabs}       
\usepackage{amsfonts}       
\usepackage{nicefrac}       
\usepackage{microtype}      
\usepackage{times}
\usepackage{epsfig}
\usepackage{graphicx}
\usepackage{amssymb}
\usepackage{times}
\usepackage{graphicx}
\usepackage{amssymb}
\usepackage[ruled,vlined]{algorithm2e}
\usepackage{blindtext}
\usepackage{times}
\usepackage{helvet}
\usepackage{courier}
\usepackage{amsmath}
\usepackage{multirow}
\usepackage{textcomp}
\usepackage{xcolor}

\title{Generative-Discriminative Variational Model for Visual Recognition}

\author{
  Chih-Kuan Yeh \\
  Research Center for IT Innovation\\
  Academic Sinica\\
  \texttt{jason6582@gmail.com} \\
  \And
   Yao-Hung Hubert Tsai \\
   Machine Learning Department\\
   Carnegie Mellon University \\
    \texttt{yaohungt@cs.cmu.edu} \\
   \AND
   Yu-Chiang Frank Wang \\
   Research Center for IT Innovation\\
   Academic Sinica\\
   \texttt{ycwang@citi.sinica.edu.tw} \\
}

\begin{document}

\maketitle
\begin{abstract}
The paradigm shift from shallow classifiers with hand-crafted features to end-to-end trainable deep learning models has shown significant improvements on supervised learning tasks. Despite the promising power of deep neural networks (DNN), how to alleviate overfitting during training has been a research topic of interest. In this paper, we present a Generative-Discriminative Variational Model (GDVM) for visual classification, in which we introduce a latent variable inferred from inputs for exhibiting generative abilities towards prediction. In other words, our GDVM casts the supervised learning task as a generative learning process, with data discrimination to be jointly exploited for improved classification. In our experiments, we consider the tasks of multi-class classification, multi-label classification, and zero-shot learning. We show that our GDVM performs favorably against the baselines or recent generative DNN models.
\end{abstract}

\section{Introduction}
\label{sec:intro}

The recent advances in deep neural networks (DNN) have shown promising results on a variety machine leaning tasks including image classification~\cite{he2016deep,krizhevsky2012imagenet}, machine translation~\cite{sutskever2014sequence} and speech recognition~\cite{hinton2012deep}. Compared to traditional learning approaches, learning DNN is to produce an over-parameterized model, which generally requires a large-scale dataset during the training stage. As a result, if the size of the training dataset is not sufficiently large, how to derive DNN models while alleviating possible overfitting problems would be of interest. For example, dropout~\cite{srivastava2014dropout} is a regularization technique which randomly removes a fraction of neurons during the training phase, and batch normalization~\cite{ioffe2015batch} imposes constraints for normalizing the produced representations. 

Another preferable strategy for alleviating overfitting is to derive latent feature representation by learning generative models. For example, Gaussian Processes Latent Variable Models (GPLVM) \cite{lawrence2003gaussian} construct low-dimensional manifolds by observing a small number of instances. More specifically, it maps latent variables with a predetermined prior into the data space via a Gaussian process. Variational autoencoders (VAE)~\cite{kingma2014auto} derives deep latent spaces by utilizing stochastic variational inference, which scales efficiently to the datasets with varying sizes. However, the above methods cannot handle supervised learning tasks. Salakhutdinov et al.~\cite{salakhutdinov2009deep} apply generative Restricted Boltzmann Machines for learning Deep Boltzmann Machines, which is latter used to fine-tune a classification network. Susskind et al.~\cite{susskind2011deep} use a gated MRF at the lowest level of deep
belief network for recognition. However, to reach the goal of supervised learning, existing generative-based approaches typically require a different (network) component which is specifically trained for introducing the discriminative ability.



In this paper, we propose a discriminative latent variable model, referred to as generative-discriminative variational model (GDVM), which can be implemented by the use of deep neural networks. We approach standard supervised learning tasks by casting discriminative learning (i.e. recognition, classification tasks) as a generative learning problem. We derive deep generative models with an intermediate latent variable, which is inferred from the input data and allows a generative process with additional discriminative capabilities. By maximizing the conditional log-likelihood via the framework of stochastic gradient variational Bayes (SGVB), our GDVM can be utilized in existing DNNs with marginally additional computation costs. We further show that our non-deterministic inference function for recognition would further maximize the margin between classes in the latent space. This is the reason why, as supported by our experiments, promising recognition results can be achieved by using our GDVM for solving classification problems. 


We highlight our contributions of the paper as follows:
\begin{itemize}

	\item We propose a generative-discriminative variational model (GDVM), which uniquely casts standard supervised recognition learning into a generative learning process, and can be easily integrated into existing DNN structures.
    \item We provide detailed derivations with fundamental supports, which explain how our model jointly exhibits generative and discriminative abilities for supervised learning. We also present insight discussions and explain how our GDVM differs from maximum-margin and existing generative DNN models.
	\item The non-deterministic inference function in our GDVM (i.e., $Q(\textbf{z}|\textbf{x})$ as detailed in Sect.~\ref{sec:3}) enforces data separation. As verified by a variety of recognition tasks (i.e., multi-class classification, multi-label classification, and zero-shot learning), improved recognition performance can be achieved.
\end{itemize}

\section{Related Work}
\label{sec:2}

{\bf Prevent Overfitting in DNN:} In order to train DNNs with sufficient generalization ability and avoid overfitting, researchers have proposed different regularizers when learning parameter weights (e.g.,~\cite{tibshirani1996regression}). More recently, a line of works in preventing overfitting in DNN is achieved by suppressing correlation between network activations~\cite{srivastava2014dropout,ioffe2015batch,cogswell2015reducing}. Alternatively, Liao et al.~\cite{liao2016learning} encourage parsimonious representations as a form of regularization. Nevertheless, the main idea of regularizing DNN is to limit the search space of the network parameters. Along with a generative variational learning framework, we regularize the distribution of the derived latent representation to fit a given simple prior, which would introduce the additional separation between classes and thus result in improved recognition performance.

{\bf Deep Generative Models:}
Generative models aim to capture data distributions so that a proper feature representation can be produced for achieving learning tasks like image inpainting~\cite{yeh2016semantic} and feature disentanglement~\cite{reed2014learning,chen2016infogan}. A number of deep generative models have been proposed and attracted the attention from the researchers, including Deep Boltzmann machine~\cite{salakhutdinov2009deep} and Deep Belief Networks~\cite{hinton2006fast}. Recently, Generative Adversarial Network (GAN)~\cite{goodfellow2014generative} and Variational Autoencoder~\cite{kingma2014auto} have shown remarkable performances on a variety of learning tasks. Recently, deep generative models can achieve promising results on recognition tasks~\cite{susskind2011deep,salakhutdinov2009deep,pu2016deep} by combining a discriminative objective function with a generative one. Our model is built upon variational inference and learning with the goal of solving classification problems. As detailed later, our model can be easily implemented via learning DNN/CNN models, which uniquely integrates and optimizes the discriminative and generative objectives.


{\bf Conditional Generative Models:} Stochastic Feed-forward Neural Network (SFNN)~\cite{tang2013learning} and Conditional Variational Autoencoder (CVAE)~\cite{sohn2015learning,walker2016uncertain} are both directed graphical models, which derive distributions of output variables based on the input data for structured output prediction (e.g., image completion or image segmentation). SFNN utilizes stochastic feedforward network with hidden layers composed of both deterministic and stochastic variables but suffers from non-trivial inferences. While CVAE utilizes Gaussian latent variables in their models, it uses different networks (with different inputs) to inference the latent variable in training and testing. The above inefficiency or inconsistency during training and testing might not be preferable in practical tasks. In this paper, our proposed variational model is learned with generative and discriminative abilities and focuses on supervised recognition tasks. In later sections, we will explain how our approach differs from existing generative models for supervised learning, and why the improved separation between data of different classes can be expected.

\section{Generative-Discriminative Variational Model (GDVM)}\label{sec:3}
\subsection{Variational Model for Supervised Learning}\label{sec:3.1}

Let $\textbf{X} =\{\textbf{x}_1,\dots,\textbf{x}_N\}$ denote a set of $N$ training instances, and $\textbf{Y} =\{\textbf{y}_1,\dots,\textbf{y}_N\}$ as the corresponding training target outputs. For supervised learning, the goal is to learn a function $\mathcal{F}$ with a loss metric $\mathcal{L}$ so that the the expected loss of $\mathcal{L}(\mathcal{F}(\textbf{X}_t), \textbf{Y}_t)$ would be minimized for given test data $\textbf{X}_t$ and their ground truth outputs $\textbf{Y}_t$. For deep learning, $\mathcal{F}$ is often parameterized by layers of neural networks $\Phi$.

In this paper, we introduce a latent variable $\textbf{z}$, which is to generatively model the conditional probability $P(\textbf{y}|\textbf{x})$ for supervised learning tasks. With the introduced latent variable $\textbf{z}$, the maximization of the above conditional probability can be written as
\begin{equation}
\label{eq:1}
P(\textbf{y}|\textbf{x}) = \int P(\textbf{y}|\textbf{x},\textbf{z}) P(\textbf{z}) d\textbf{z},
\end{equation}
where $P(\textbf{y}|\textbf{x},\textbf{z})$ can be calculated by a neural network function $\Phi(\textbf{x},\textbf{z})$. We note that, if $\textbf{y}$ is unknown, $\textbf{z}$ is assumed to be sampled from a given prior independent of the data distribution for $\textbf{x}$. In the context of this work, we have this prior as $\mathcal{N}(0;I)$ for simplicity.

\begin{figure}[t!]
	\centering
	\vspace{-5mm}
	\includegraphics[width=0.99\textwidth]{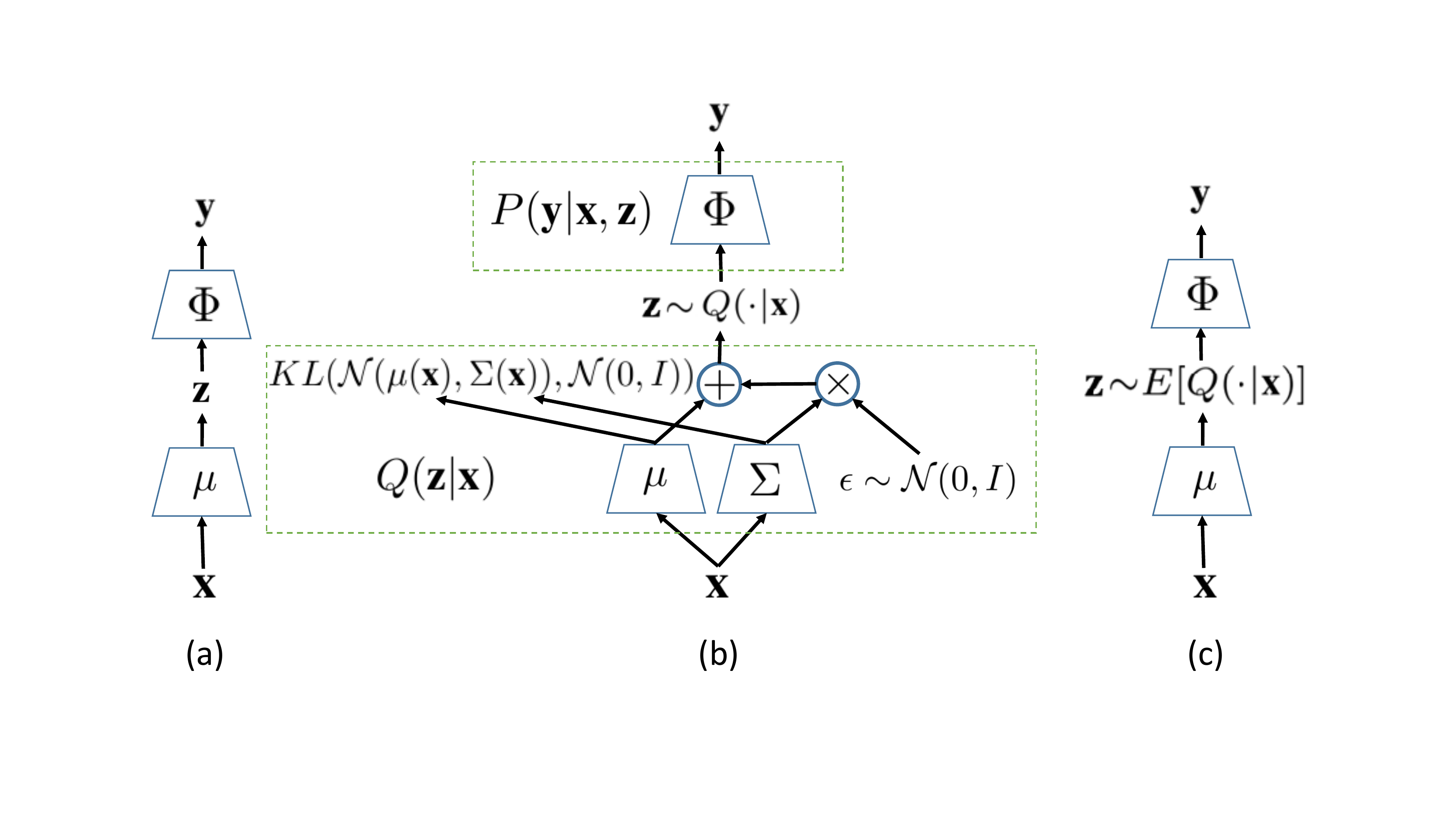}
	\vspace{-16mm}
	\caption{The architecture of generative-discriminative variational models (with networks $\Phi$, $\mu$ and $\Sigma$): (a) standard deep learning architecture mapping input $\textbf{x}$ to output $\textbf{y}$ with a latent layer $\textbf{z}$, (b) learning of GDVM with probability estimation functions $P(\textbf{y}|\textbf{x}, \textbf{z})$ and $Q(\textbf{z}|\textbf{x})$, and (c) prediction using GDVM.}
	\vspace{-2mm}
	\label{fig:structure}
\end{figure}




Unfortunately, the direct maximization of~\eqref{eq:1} requires the sampling of a large number of $\textbf{z}$ from the given prior, which would not be computationally efficient. Thus, as suggested in~\cite{kingma2014auto,doersch2016tutorial}, we wish to perform such sampling while satisfying ~\eqref{eq:1}. That is, we choose to maximize~\eqref{eq:1} with~$\textbf{z}$ which makes $P(\textbf{y}|\textbf{x},\textbf{z})$ not negligible. To be more precise, we aim at estimating $P(\textbf{z}|\textbf{x},\textbf{y})$ for achieving the above goal. For the task of supervised learning, the input data $\textbf{x}$ would contain the target information of $\textbf{y}$. Thus, we allow a function $Q(\textbf{z}|\textbf{x})$ which approximates the intractable posterior outputs $P(\textbf{z}|\textbf{x},\textbf{y})$. 

\subsection{The Variational Bound}\label{sec:3.2}

With the introduced latent variable $\textbf{z}$, we now relate $P(\textbf{y}|\textbf{x})$ to $\textbf{E}_{\textbf{z}\in Q}P(\textbf{y}|\textbf{x},\textbf{z})$, which is the key of variational Bayesian methods. To start, we derive the Kullback-Leiber divergence (KL-divergence, abbreviated as KL) between $Q(\textbf{z}|\textbf{x})$ and $P(\textbf{z}|\textbf{x},\textbf{y})$ as follows: \begin{equation}\label{eq:2}{KL}\Big(Q(\textbf{z}|\textbf{x})||P(\textbf{z}|\textbf{x},\textbf{y})\Big) = \textbf{E}_{z \sim Q(\cdot |\textbf{x})}[\log Q(\textbf{z}|\textbf{x}) - \log P(\textbf{z}|\textbf{x},\textbf{y})].\end{equation}
By applying the Bayes rule to $P(\textbf{z}|\textbf{x},\textbf{y})$, we have the following equation containing $\textbf{E}_{\textbf{z}\in Q}P(\textbf{y}|\textbf{x},\textbf{z})$ and $P(\textbf{y}|\textbf{x})$:
\begin{equation}\label{eq:3} \log P(\textbf{y}|\textbf{x}) - {KL}\Big(Q(\textbf{z}|\textbf{x})||P(\textbf{z}|\textbf{x},\textbf{y})\Big) = \textbf{E}_{\textbf{z} \sim Q(\cdot |\textbf{x})}[ \log P(\textbf{y}|\textbf{x},\textbf{z} )] -
{KL}\Big(Q(\textbf{z}|\textbf{x})||P(\textbf{z}|\textbf{x})\Big).\end{equation}


Due to the non-negativity of KL-divergence, we can effectively view the right hand side in~\eqref{eq:3} as the variational lower bound of $P(\textbf{y}|\textbf{x})$. By maximizing this lower bound, the maximization of $P(\textbf{y}|\textbf{x})$ and minimization of ${KL}\Big(Q(\textbf{z}|\textbf{x})||P(\textbf{z}|\textbf{x},\textbf{y})\Big)$ would be equivalently and implicitly achieved. The former meets our goal of supervised learning, while the latter allows us to obtain an accurate estimation for the intractable term $P(\textbf{z}|\textbf{x},\textbf{y})$ with $Q(\textbf{z}|\textbf{x})$. In our proposed model, we perform stochastic gradient descent on the right hand side of~\eqref{eq:3} and the optimization process will be discussed in the later sections.

\subsection{Optimizing The Variational Bound} \label{sec:3.3}

As noted above, optimizing the variational bound of $P(\textbf{y}|\textbf{x})$ is to maximize~\eqref{eq:3}. More specifically, this is realized by the maximization of the RHS of~\eqref{eq:3}, with additional insights as we now discuss.

Let $Q(\cdot)$ and $P(\cdot)$ as two neural networks, and $\textbf{z}$ as the hidden layer. As illustrated in Figure~\ref{fig:structure}, we approach the maximization of $\textbf{E}_{\textbf{z} \sim Q(\cdot |\textbf{x})}[ \log P(\textbf{y}|\textbf{x},\textbf{z} )]$ in~\eqref{eq:3} by optimizing this network structure. On the other hand, minimization of ${KL}\Big(Q(\textbf{z}|\textbf{x})||P(\textbf{z}|\textbf{x})\Big)$ in~\eqref{eq:3} can be regarded as regularizing the hidden layer of $\textbf{z}$, which enforces the output distribution of $Q(\textbf{z}|\textbf{x})$ to fit a given prior (e.g., $ \mathcal{N}(0,I)$ in our work). With networks $Q(\cdot)$ and $P(\cdot)$ properly selected and optimized, we can therefore maximize the variational lower bound of $P(\textbf{y}|\textbf{x})$ accordingly.

In Figure~\ref{fig:structure}, we particularly have the output distribution of $Q(\cdot)$ to be $\mathcal{N}(\mu(\textbf{x}),\Sigma(\textbf{x}))$, where $\mu(\cdot)$ and $\Sigma(\cdot)$ are two neural network components. With this design, the above regularization of $\textbf{z}$ can be integrated easily as follows. Recall that $\textbf{z}$ is sampled from the prior distribution of $\mathcal{N}(0,I)$, which is independent of $\textbf{x}$ when $\textbf{y}$ is unknown. Thus, the output distribution of $P(\textbf{z}|\textbf{x})$ would be $\mathcal{N}(0,I)$, and the KL-Divergence between $Q(\textbf{z}| \textbf{x})$ and $P(\textbf{z}|\textbf{x})$ can be computed as:
\begin{equation}\label{eq:4} {KL}\Big(\mathcal{N}(\mu(\textbf{x}),\Sigma(\textbf{x}))||\mathcal{N}(0,I)\Big) = \frac{1}{2}[ \mathbf{tr}(\Sigma(\textbf{x})) + (\mu(\textbf{x}))^\top(\mu(\textbf{x})) - k -\log \det(\Sigma(\textbf{x}))], \end{equation}
where $k$ is the dimensionality of $\textbf{z}$. With a properly designed function $Q$ (as in our GDVM), minimization of ${KL}\Big(Q(\textbf{z}|\textbf{x})||P(\textbf{z}|\textbf{x})\Big)$ can be simply achieved by applying stochastic gradient decent.

Another advantage of using our network design $Q$ is that, maximization of $\textbf{E}_{\textbf{z} \sim Q(\cdot |\textbf{x})}[ \log P(\textbf{y}|\textbf{x},\textbf{z} )]$ can also be realized by using neural network techniques. Without advancing the network structure like ours, the dependence of $\textbf{E}_{\textbf{z} \sim Q(\cdot |\textbf{x})}[ \log P(\textbf{y}|\textbf{x},\textbf{z} )]$ on $Q$ cannot be observed after sampling $\textbf{z}$ from an arbitrarily designed $Q$. This is the reason why we apply the reparameterization trick \cite{kingma2014auto} in constructing network components in $Q$, which relates the output variable $\textbf{z}$ to $\mathcal{N}(\mu(\textbf{x}),\Sigma(\textbf{x}))$ and maintains its dependency on $Q$. As a result, stochastic gradient descent can be directly applied to update $\textbf{E}_{\textbf{z} \sim Q(\cdot |\textbf{x})}[ \log P(\textbf{y}|\textbf{x},\textbf{z})]$.

\begin{figure}[t!]
	\centering
	\includegraphics[width=0.99\textwidth]{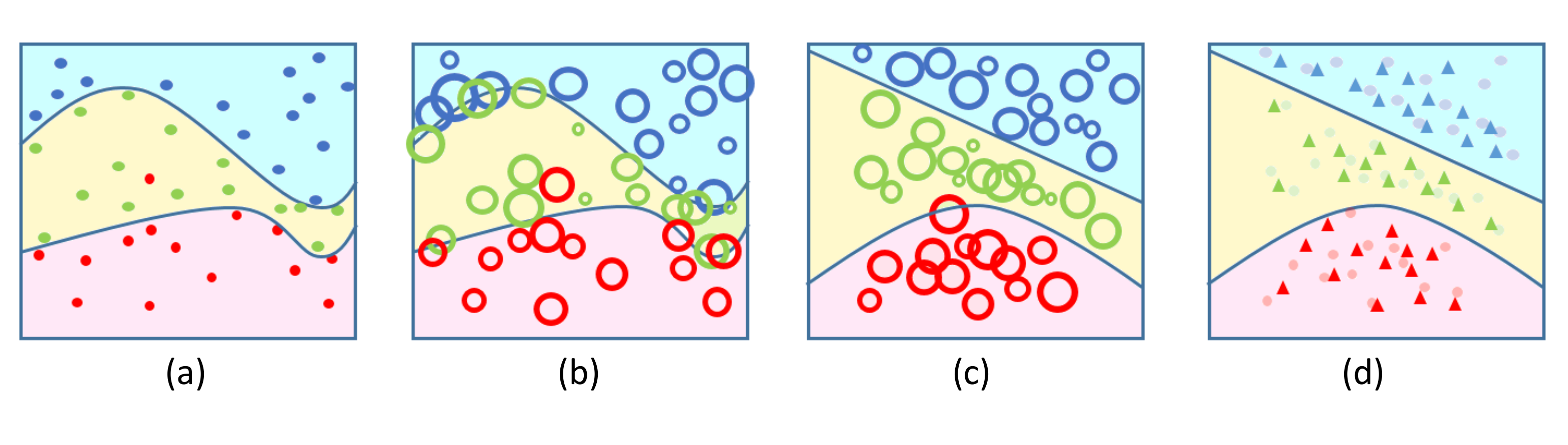}
	\vspace{-6mm}
	\caption{Generative-discriminative variational models for classification. The dots/circles in different colors depict instances of different classes. (a) classifiers learned by standard CNNs, (b) classifiers of (a) cannot be generalized to our variational model with $\mu$ and $\Sigma$ for describing data via latent representation $\textbf{z}$, (c) our model updates $\mu$, $\Sigma$, $\textbf{z}$ via back propagating the classification loss, which results in tne improved data separation, and (d) prediction of test data (i.e., triangles) without $\Sigma$.}
	\vspace{-1mm}
	\label{fig:corr}
\end{figure}

While we show that our design of $Q(\cdot)$ makes maximizing the variational bound of $P(\textbf{y}|\textbf{x})$ tractable, we further explain this process via a representation learning aspect of view. As depicted in Figure~\ref{fig:structure}, $\mu(\textbf{x})$ can be seen as a standard neural network mapping from the input $\textbf{x}$ to the latent variable $\textbf{z}$, and $\Sigma(\textbf{x})$ can be seen as the observed data variation. Take recognition tasks for example, by maximizing $\textbf{E}_{\textbf{z} \sim Q(\cdot |\textbf{x})}[ \log P(\textbf{y}|\textbf{x},\textbf{z})]$, our goal is to achieve satisfactory recognition performances with outputs of $\Sigma(\cdot)$ imposed on the latent space.

We note that, if two input instances with distinct labels result in similar representations in the latent space (i.e., similar $\mu(\textbf{x})$), it is possible to learn a complex classification model for separating one from another. However, with the introduction of $\Sigma(\textbf{x})$ in the latent representation, it is less likely to have the same model to exhibit comparable discriminating ability. As a result, the loss observed at the output of the network (e.g., Figure~~\ref{fig:structure}(b)) would be propagated to update both $\mu(\cdot)$ and $\Sigma(\cdot)$. As illustrated in Figure \ref{fig:corr}, this allows the learning of latent representation \textbf{z} for a improved recognition performance.


As for calculating $P(\textbf{y}|\textbf{x},\textbf{z})$ in~\eqref{eq:3}, we also advance a neural network $\Phi(\cdot)$ as depicted in Figure~\ref{fig:structure}(b). In the context of this work, the learning of this network is achieved by optimizing $\textbf{E}_{\textbf{z} \sim Q(\cdot |\textbf{x})}[ P(\textbf{y}|\textbf{x},\textbf{z})]$
It is worth noting that, the input $\textbf{z}$ of the network $\Phi$ is sampled from the encoding function/network $Q(\cdot |\textbf{x})$ with input $\textbf{x}$. Thus, we simply have $\Phi(\textbf{x},\textbf{z})$ = $\Phi(\textbf{z})$. More precisely, we let the neural network $\Phi(\textbf{z})$ learned from $\textbf{z} \sim Q(\cdot |\textbf{x})$, and $\textbf{E}_{\textbf{z} \sim Q(\cdot |\textbf{x})}[ P(\textbf{y}|\textbf{x},\textbf{z})]$ can be calculated accordingly.

Given training data $\textbf{X}$ and $\textbf{Y}$ with the proposed network structure consisting of $\Phi$, $\mu$, and $\Sigma$, we now summarize the overall loss function as:


\begin{equation}\label{eq:5}
\mathcal{L}_{all} = \sum_{i = 1}^{N}\{\mathcal{L}
(\textbf{y}_i,\textbf{E}_{\textbf{z} \sim Q(\cdot |\textbf{x}_i)}[ \Phi(\textbf{z})])+
\beta \{\frac{1 }{2}[ k +\log \det(\Sigma(\textbf{x}_i))]- \mathbf{tr}(\Sigma(\textbf{x}_i)) - \mu(\textbf{x}_i)^\top\mu(\textbf{x}_i)\}\} ,
\end{equation}
Note that the first term calculates the loss function for classification. As noted in~\eqref{eq:4}, the second term enforces the distribution of $Q(\textbf{z}|\textbf{x})$ to fit $\mathcal{N}(0,I)$. Selected via validation, $\beta$ is a parameter balancing between the two terms in~\eqref{eq:3}. With~\eqref{eq:5} calculated, we can update the network via stochastic back propagation. Algorithm~\ref{alg:learning} summaries the learning process of our proposed method.

Once the learning of the above network architecture is complete, prediction of the output $\textbf{y}$ for each test input $\textbf{x}_i$ can be performed via the following deterministic inference process\footnote{Alternatively, we can draw multiple $\textbf{z}$ from the prior distribution $\mathcal{N}(0;I)$ and use the average of $P(\textbf{y}|\textbf{x},\textbf{z})$ as prediction.}:
\begin{equation}\label{eq:6}
\textbf{y}_i = \arg \max_\textbf{y} P(\textbf{y}|\textbf{x}_i,\textbf{z}), \textbf{z} = \textbf{E}_{\textbf{z}\sim Q(\cdot |\textbf{x}_i)}[\textbf{z}],
\end{equation}
where $\textbf{E}_{\textbf{z}\sim Q(\cdot |\textbf{x}_i)}[\textbf{z}]$ is computed as $\mu(\textbf{x}_i)$. 

\begin{algorithm}[t]
	\label{alg:learning}
	\DontPrintSemicolon
	\caption{Learning of Generative-Discriminative Variational Model}
	\KwIn{Training set $\textbf{D}=\{(\textbf{x}_i,\textbf{y}_i)\}_{i=1}^N$}
	Randomly initialize neural network $\Phi, \mu, \Sigma$.
	\\
	\Repeat{Converge}
	{
		Randomly select a batch of instances $\textbf{x}_j$ and $\textbf{y}_j$ \\
		Sample $\mathbf{\epsilon}_j$ from {$\mathcal{N}(0,I)$ } \\
		Let $\textbf{z}_j = \mathbf{\epsilon}_j \times \Sigma(\textbf{x}_j) + \mu(\textbf{x}_j) $\\
		Minimize the loss function in~\eqref{eq:5}\\
		Update $\Phi, \mu, \Sigma$ via back propagation	
	}
    \KwOut{Network parameters $\Phi, \mu, \Sigma$}
\end{algorithm}


\subsection{Additional Remarks}\label{sec:3.4}

{\bf Comparison to Variational Models:}
Our proposed variational learning architecture can be viewed as a deep directed graphical model with Gaussian latent variables as Kingma et al.~\cite{kingma2014auto} did. In~\cite{kingma2014auto}, Variational Aotoencoder (VAE) is designed to model latent feature representations in an unsupervised learning setting. To generalize to supervised learning settings, Conditional Variational Autoencoder (CVAE)~\cite{sohn2015learning} is a conditional directed graphical model, whose inputs $\textbf{x}$ modulate a Gaussian prior on latent variables $\textbf{z}$ for producing the outputs $\textbf{y}$. When training CVAE, a recognition network $\theta_r(\textbf{z}|\textbf{x},\textbf{y})$ is learned to sample $\textbf{z}$ for deriving the generative network $\theta_g(\textbf{y}|\textbf{x},\textbf{z})$. On the other hand, the prediction stage of CVAE requires a (different) prior network $\theta_p(\textbf{z}|\textbf{x})$ for producing $\textbf{z}$. However, the networks utilized for CVAE training and prediction are different (i.e., $\theta_r$ and $\theta_g$ vs. $\theta_p$ and $\theta_g$), which might not be optimal.

Extended from CVAE, Gaussian stochastic neural network (GSNN)~\cite{sohn2015learning} applies the same network architecture for $\theta_r$ and $\theta_p$. Although this makes the training and prediction process consistent, such design might result in deterministic $\textbf{z}$, which would degrade the model into a non-variational one. In this paper, we advance a unified variational model for supervised learning, in which we fit the distribution of $\textbf{z}$ as $\mathcal{N}(0,I)$ when minimizing the classification loss (i.e.,~\eqref{eq:5}). Later in experiments, we will compare our results with those produced by GSNN for supporting the above remark.



{\bf Comparison to Maximum-Margin Methods:}
In a representation learning point of view, our method resembles approaches with maximum margin criteria (MMC), which share the same goal of increasing data separation. However, most MMC-based approaches~\cite{taskar2004max,cortes1995support,hariharan2010large} aim at maximizing a pre-determined margin of data from different classes. Instead, our model enforces the separation of nearby instances (i.e., $\mu(\textbf{x})$) with distinct labels without the need to explicitly describe the margin. As explained in Section~\ref{sec:3.3}, this is achieved by back-propagating the loss of~\eqref{eq:5} to update our network architecture, which allows proper $\mu(\textbf{x})$ and $\Sigma(\textbf{x})$ for describing the observed data.

{\bf Comparison to Noise-Imposing Learning Methods:}
We note that, adding noise to a hidden layer has previously been seen in Dropout~\cite{srivastava2014dropout}, while denoising autoencoder~\cite{vincent2010stacked} and data augmentation methods~\cite{krizhevsky2012imagenet} can be considered as imposing noisy/variation information in the input layer. As mentioned in Section~\ref{sec:3.3}, our introduction of $\Sigma(\textbf{x})$ in $Q(\textbf{z}| \textbf{x})$ can be viewed as adding data-dependent variation information into the derived latent representation $\textbf{z}$. Such information is learned by our proposed network architecture without the prior knowledge of the data. In the following section, we will also provide quantitative evaluation and comparisons to confirm the effectiveness of our method.

\section{Experiments}\label{sec:4}

For evaluating the performance of our proposed GDVM, we conduct experiments on multi-class classification, multi-label classification, and zero-shot learning. We compare our method to a baseline CNN as well as GSNN~\cite{sohn2015learning}, which is implemented via setting $\beta = 0$ in~\eqref{eq:5} without recurrent connection. In our experiments, we use the same network structures (i.e.,  networks $\Phi$ and $\mu$) for the baseline CNN, GSNN, and our proposed model, and perform parameter selection via validation using $20\%$ of the training data.


\subsection{Multi-Class Classification}\label{sec:4.1}

We consider the benchmark dataset of CIFAR10~\cite{krizhevsky2009learning} for multi-class classification\footnote{Due to space limit, experiments on the MNIST dataset are presented in the supplementary.}. CIFAR10 consists of 60,000 32×32 images with 10 categories. We use 50,000 images across all categories for training and the remaining ones for testing. For the baseline CNN architecture, we build a network based on VGG~\cite{simonyan2014very} with fewer convolution and fully connected neurons due to the small size of images in CIFAR10. We choose the dimension for $\textbf{z}$ as 64. The detailed network structure is presented in the supplementary, which is learned via SGD optimizer with learning rate 0.05 and momentum 0.9. We perform validation on the total epochs in the range of $[200, 300, 400]$, and select $\beta$ in ~\eqref{eq:5} from the range of $[0.1, 0.5, 1.0]$. Finally, we report average accuracy of 5 runs for evaluation and comparison.

\begin{table}[]
	\centering
	\def\arraystretch{1.2}
	\begin{tabular}{l||l|l|l|l}
		\hline
		$\#$ of Training Instances            & 5000            & 10000           & 20000           & 50000           \\ \hline\hline
		CNN w/o Dropout      & $46.63 \pm 0.54$ & $52.63 \pm 1.10$ & $60.64 \pm 0.88$ & $71.75 \pm 0.75$ \\ \hline
		GSNN w/o Dropout  & $47.04 \pm 1.24$ & $52.70 \pm 0.76$ & $60.20 \pm 1.39$ & $71.17 \pm 1.68$ \\ \hline
		Ours w/o Dropout  & $56.30 \pm 0.24$ & $61.44 \pm 1.08$ & $68.23 \pm 1.04$ & $76.40 \pm 0.32$ \\ \hline\hline
		CNN w/ Dropout   & $49.42 \pm 0.45$ & $56.95 \pm 0.83$ & $66.39 \pm 0.98$ & $77.88 \pm 0.52$ \\ \hline
		GSNN w/ Dropout  & $48.51 \pm 1.37$ & $55.97 \pm 0.86$ & $ 65.76 \pm 1.88$ & $76.35 \pm 1.36$ \\ \hline
		Ours w/ Dropout & $\textbf{60.43} \pm 1.33$  & $\textbf{68.30} \pm 1.18$ & $\textbf{76.05} \pm 0.39$  & $\textbf{83.39} \pm 0.28$ \\ \hline
	\end{tabular}
	\vspace{2mm}
	\caption{Classification results on CIFAR10. For each setting, the best performance is shown in bold.}
	\label{table:cifar}
\end{table}

\begin{figure}[t!]
	\centering
	\vspace{-2mm}
	\includegraphics[width=0.95\textwidth]{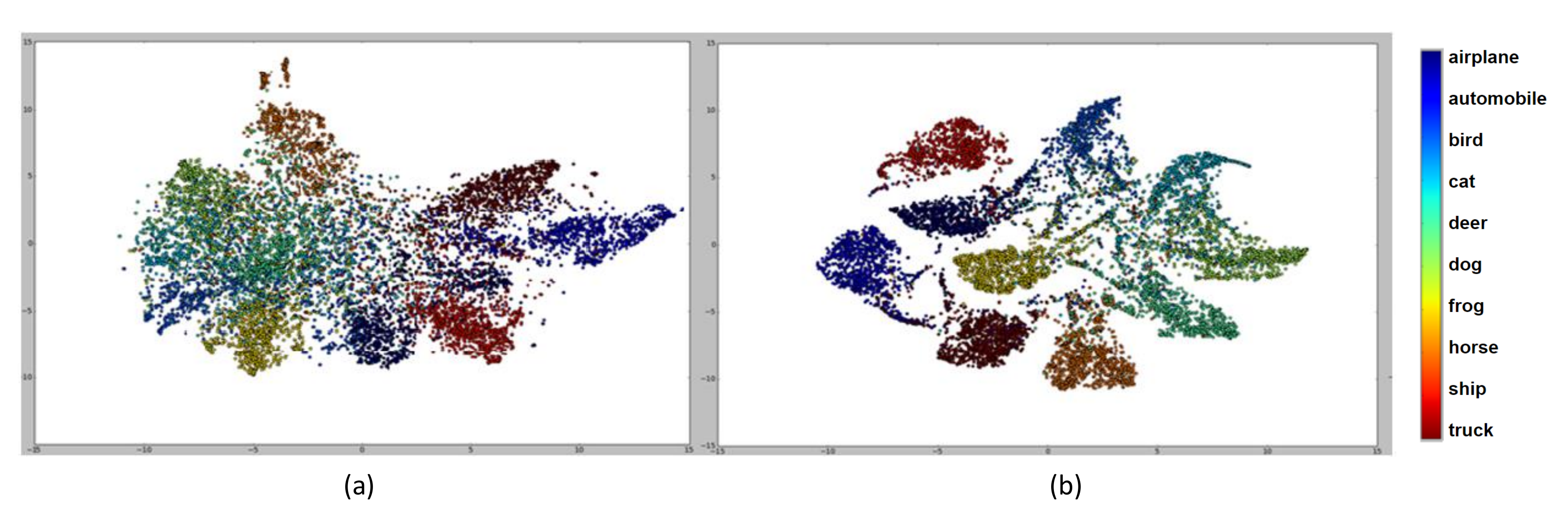}
\vspace{-2mm}	
\caption{t-SNE visualization of the latent spaces derived by (a) standard CNN and (b) our GDVM on CIFAR10. Note that the improved separation between data of different classes can be observed in (b), while the average recognition rates for (a) and (b) are $77.88\%$ and $83.39\%$, respectively. }
	
	\label{fig:visual}
\end{figure}

To assess the ability of our GDVM in alleviating overfitting, we perform experiments using $[5000, 10000, 20000, 50000]$ training instances. In Table~\ref{table:cifar}, we observe that our approach performed favorably against the baseline CNN model with or without dropout. When the size of training data is $20,000$, we further observe a $10\%$ improvement (e.g., $76.05$ vs. $66.39$). This supports the use of our model for achieving satisfactory performance given a limited amount of training data. It can also be seen that GSNN did not necessarily produce improved performance over the baseline CNN. This implies that, without regularizing the latent representation like ours (see~\eqref{eq:5}), GSNN might not exhibit sufficient generative capability especially with small training data sizes. Figure~\ref{fig:visual} visualizes the latent spaces derived by standard CNN and our model. From this figure, it is clear that our method resulted in improved margin for separating data of different classes.

\subsection{Multi-Label Classification on NUS-WIDE LITE} \label{sec:4.2}
We now evaluate the performance of our GDVM on multi-label classification. Different from multi-class classification, multi-label classification requires the prediction of multiple labels to an input instance. With the proposed generative-discriminative variational model, we aim at verifying that the derived latent representation would exhibit promising ability in describing data with multiple labels, while the difference between data can still be properly identified.

To perform the experiments, we consider the multi-label dataset of NUS-WIDE LITE~\cite{chua2009nus}, which includes 55,615 images with a total of 81 concepts (i.e., labels). Given the pre-specified training and test sets of this dataset, we apply convolution layers of Alexnet pre-trained on ImageNet as the network components for all methods of interest, and Binary Cross Entropy is utilized for calculating the loss function. The detailed network structure is presented in the supplementary. An Adam optimizer with learning rate 0.001 is used, and the number of training epochs is set to 20. We perform validation for $\beta$ in the range of $[0.1, 0.5, 1.0, 5.0, 10.0]$. For evaluation and comparison, we report the average of results (in terms of Micro-F1 and Macro-F1) on test data with 10 runs. The results are listed in Table \ref{table:nuswide_lite}, which confirm the effectiveness of our GDVM, which performed favorably against standard CNN and GSNN. The lack of latent space regularization of GSNN again was not able to achieve comparable results to those reported by CNN, which not only implies the possible non-variational model was produced by GSNN but also supports the use of our proposed network architecture for exhibiting both generative and discriminative abilities.

\subsection{Zero-Shot Learning} \label{sec:4.3}

Finally, we apply our GDVM for solving a more challenging task of zero-shot learning, which requires one to recognize instances of classes that are unseen during training. We take the AWA dataset~\cite{lampert2014attribute}, and follow the same data split as did in~\cite{akata2015evaluation}. We choose Cross Modal Transfer (CMT)~\cite{socher2013zero} as the baseline CNN model for zero-shot learning, which performs recognition using the semantic space derived from training data of seen classes.

When training our GDVM for zero-shot learning, we view the input images as $\textbf{x}$ and the output $\textbf{y}$ as the semantic vector of the associated class. In other words, with target outputs $\textbf{y}$ as semantic vectors of interest, we approach zero-shot learning by solving a regression task using our proposed model. The convolution layers in the network structures are those of pre-trained GoogLeNet, with detailed network structures presented in the supplementary.

As for the validation data, it is applied for choosing the total number epochs, as well as $\beta$ in~\eqref{eq:5} between $[0.0001, 0.001.0.01, 0.1]$. Note that the $\beta$ value is expected to be smaller since L2-distance is used for loss calculation in CMT rather than binary-cross-entropy as used in previous CNN models. We report the average of top-1 accuracy for 10 runs, and present the results in Table~\ref{table:AWA}. From this table, it is clear that CMT with the introduction of our GDVM was able to achieve the best results among all CNN-based methods. This confirms the effectiveness of the derived latent space for describing and discriminating between (attribute) data, which would be preferable for the challenging task of zero-shot learning.

\begin{table}[]
	\centering
	\def\arraystretch{1.2}
	\vspace{2mm}
	\begin{tabular}{l||l|l}
		\hline
          & Micro-F1           & Macro-F1                   \\ \hline\hline
		CNN  & $58.262 \pm 3.289$ & $38.012 \pm 0.825$ \\ \hline

		GSNN & $51.194 \pm 2.363$ & $36.063 \pm 0.801$  \\ \hline
		Ours & $\textbf{62.378} \pm 0.102$ & $\textbf{38.532} \pm 0.620$  \\ \hline
\hline
	\end{tabular}
	\vspace{2mm}
	\caption{Performance of multi-label classification on NUS-WIDE LITE.}
	\label{table:nuswide_lite}
\end{table}

\begin{table}[]
	\centering
	\def\arraystretch{1.2}
	\vspace{2mm}
	\begin{tabular}{l|l}
		\hline
		& Accuracy                        \\ \hline\hline
		CMT  & $58.413 \pm 0.848$  \\ \hline
		CMT + GSNN & $60.155 \pm 2.589$   \\ \hline
		CMT + Ours & $\textbf{66.532} \pm 2.609$   \\ \hline
	\end{tabular}
	\vspace{2mm}
	\caption{Performance of zero-shot learning on the AWA dataset}
	\label{table:AWA}
\end{table}

\subsection{Computational Cost} \label{sec:4.4}
As shown in Figure~\ref{fig:structure}, the difference between the network architecture of our GDVM and that of standard DNN is that we introduce an additional layer $\Sigma(\textbf{x})$ with an operation layer $\mathcal{N}(\mu(\textbf{x}),\Sigma(\textbf{x})) = \mu(\textbf{x}) + \epsilon \times \Sigma(\textbf{x})$. Since $\Sigma(\textbf{x})$ shares all layers except for the highest one with $\mu(\textbf{x})$, the additional cost for training such network components would be marginal. For quantitative comparisons, we train a neural network with a mini-batch size of 100 on CIFAR10 and evaluate the results over 100 epochs. For the baseline CNN with dropout, we observed the training time per image as $296.52$ microseconds, while ours (also with dropout) with $299.22$ microseconds. The computation time estimates were performed on a GTX 1080 GPU. Therefore, we confirm that our GDVM was able to achieve promising classification performance without significantly sacrificing the training time.


\section{Conclusion}\label{sec:5}

In this paper, we presented a novel generative-discriminative variational model (GDVM), which uniquely advances a generative model with a deterministic discriminative objective for supervised learning. Our GDVM is implemented via stochastic neural networks with Gaussian latent variables, while the network architecture can be easily built upon existing CNN structures with marginal computational costs. We discussed the differences between our GDVM and existing generative or noise-imposing DNN models, and we explained why GDVM could be viewed as maximum-margin models without the need to explicitly determine the margin. As verified by a variety of classification tasks, our GDVM was shown to be preferable over baseline or state-of-the-art DNN models, especially when the training set size was not sufficiently large.


%
%
\bibliographystyle{ieee}
\bibliography{nips2017_ref}

\section{Supplement: Multi-Class Classification on MNIST} \label{sec:s2}
We perform experiments on multi-class classification using the hand-written digit dataset of MNIST, which consists of 60,000 images of 10 categories. Following the data split of~\cite{krizhevsky2009learning}, we apply the CNN architecture for training our GDVM (see Sect. \ref{sec:s1} for detailed structure), and utilize the RMSProp optimizer with the learning rate of 0.001 and $\rho$ of 0.9. With 20\% of training data as the validate set, we select $\beta$ from $[0.1, 0.5, 1.0, 5.0, 10.0]$ on the total epochs in the range of $[25, 50, 100, 150]$. Finally, we report the average accuracy on test data of 10 runs.

{\bf Classification Results:}

We perform experiments with $[500, 1000, 2000, 4000]$ training instances. Throughout the classification experiments, we fix the dimension for latent layer $\textbf{z}$ as 64. Table \ref{table:mnist} lists and compares the results. From this table, we see that our GDVM generally achieved improved or comparable results, compared to standard CNN and GSNN. We note that, nevertheless, all DNN-based methods with dropout were able to achieve the average accuracy of $98.88\%$ with the assess of full training data set of MNIST.

\begin{table}[ht]
	\centering
	\def\arraystretch{1.2}
	\begin{tabular}{l||l|l|l|l}
		\hline
		Training Instances            & 500            & 1000           & 2000           & 4000           \\ \hline\hline
		CNN w/o Dropout      & $86.03 \pm 0.99$ & $90.57 \pm 2.54$ & $94.35 \pm 1.89$ & $96.12 \pm 1.21$ \\ \hline
		GSNN w/o Dropout  & $86.30 \pm 0.73$ & $92.22 \pm 1.35$ & $94.45 \pm 1.08$ & $96.03 \pm 1.05$ \\ \hline
		Ours w/o Dropout  & $89.75 \pm 1.70$ & $93.90 \pm 0.93$ & $\textbf{95.62} \pm 1.28$ & $96.71 \pm 0.43$ \\ \hline\hline
		CNN w/ Dropout   & $86.36 \pm 3.48$ & $91.52 \pm 2.93$ & $93.87 \pm 2.11$ & $95.53 \pm 1.49$ \\ \hline
		GSNN w/ Dropout  & $89.06 \pm 1.11$ & $92.89 \pm 1.17$ & $ 95.32 \pm 1.47$ & $96.50 \pm 0.58$ \\ \hline
		Ours w/ Dropout & $\textbf{90.56} \pm 1.30$  & $\textbf{94.02} \pm 1.41$ & $95.47 \pm 1.02$  & $\textbf{96.88} \pm 0.73$ \\ \hline
	\end{tabular}
	\vspace{2mm}
	\caption{Performance comparisons on MNIST.}
	\label{table:mnist}
\end{table}

\begin{figure}[t!]
	\centering
	\vspace{-5mm}
	\includegraphics[width=0.98\textwidth]{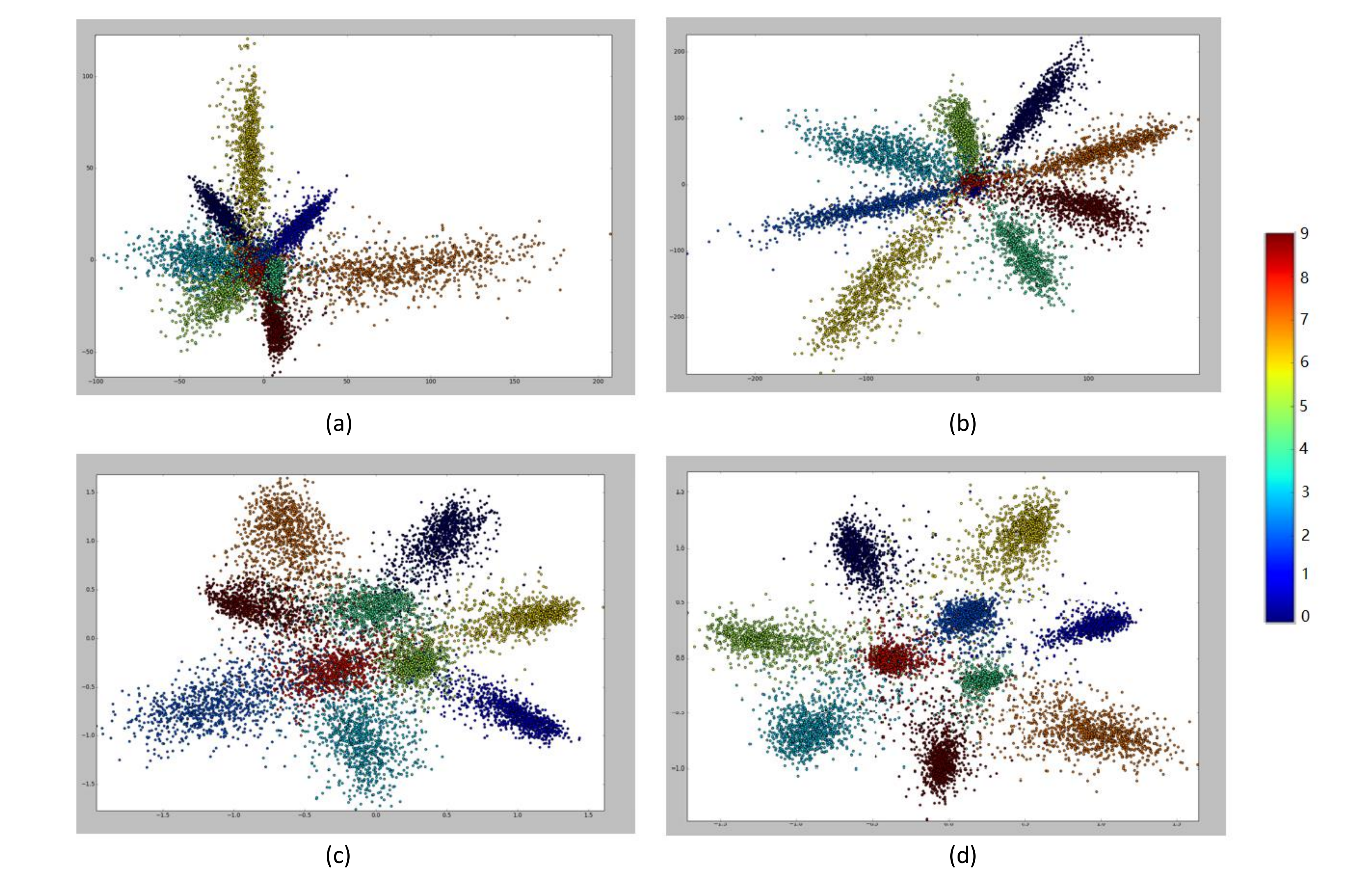}
	\caption{Visualization of the data distributions of the latent representation of MNIST data. Results of baseline CNN trained on $1000$ and $4000$ instances are shown in (a) and (b), while those of our method are shown in (c) and (d), respectively. Different colors denote class labels.}
	\label{fig:visual_mnist}
\end{figure}

{\bf Visualization Results:}
We additionally show visualization results for our model. Since the image in MNIST has a lower resolution of 28 $\times$ 28 pixels, we set the dimension of the latent layer $\textbf{z}$ as 2 for visualization purpose. Thus, without the need to apply dimension reduction technique like t-SNE, we are able to observe the data distribution of $\mu(\textbf{x})$ directly. As illustrated in Fig. \ref{fig:visual_mnist}, the latent layer $\mu(\textbf{x})$ of our GDVM resulted in better separation between classes, while that of standard CNN was not as satisfactory. This supports the above quantitative evaluation, and verifies the use of our GVDM would be preferable for classification.

\section{Supplement: Details of The Network Architectures}
\label{sec:s1}
We now present the network architecture of our generative-discriminative variational model. The network architectures for multi-class classification on CIFAR10 are shown in Tables~\ref{mu_cifar} and~\ref{phi_cifar}. Those of MNIST are shown in Tables~\ref{mu_mnist} and~\ref{phi_mnist}. For multi-label classification on NUSWIDE LITE, Tables~\ref{mu_nuswide} and~\ref{phi_nuswide} list our network designs, and those for zero-shot learning on AWA are shown in Tables~\ref{mu_awa} and~\ref{phi_awa}. For the experiments using DNN models without dropout, we simply remove all dropout layers when learning networks.

\begin{table}[h]
	\centering
	\begin{tabular}{l||l|l|l|l}
		\hline
		layer              & op.     & size-in  & size-out & kernel/dropout rate \\ \hline\hline
		\multirow{3}{*}{1} & conv    & 3$\times$32$\times$32  & 32$\times$32$\times$32 & 2$\times$2                 \\ \cline{2-5}
		& relu    & 32$\times$32$\times$32 & 32$\times$32$\times$32 & -                   \\ \cline{2-5}
		& dropout & 32$\times$32$\times$32 & 32$\times$32$\times$32 & 0.2                 \\ \hline
		\multirow{3}{*}{2} & conv    & 32$\times$32$\times$32 & 32$\times$32$\times$32 & 3$\times$3                 \\ \cline{2-5}
		& relu    & 32$\times$32$\times$32 & 32$\times$32$\times$32 & -                   \\ \cline{2-5}
		& pool    & 32$\times$32$\times$32 & 32$\times$16$\times$16 & 2$\times$2                 \\ \hline
		\multirow{3}{*}{3} & conv    & 32$\times$16$\times$16 & 64$\times$16$\times$16 & 3$\times$3                 \\ \cline{2-5}
		& relu    & 64$\times$16$\times$16 & 64$\times$16$\times$16 & -                   \\ \cline{2-5}
		& dropout & 64$\times$16$\times$16 & 64$\times$16$\times$16 & 0.2                 \\ \hline
		\multirow{3}{*}{4} & conv    & 64$\times$16$\times$16 & 64$\times$16$\times$16 & 3$\times$3                 \\ \cline{2-5}
		& relu    & 64$\times$16$\times$16 & 64$\times$16$\times$16 & -                   \\ \cline{2-5}
		& pool    & 64$\times$16$\times$16 & 64$\times$8$\times$8   & 2$\times$2                 \\ \hline
		\multirow{3}{*}{5} & conv    & 64$\times$8$\times$8   & 128$\times$8$\times$8  & 3$\times$3                 \\ \cline{2-5}
		& relu    & 128$\times$8$\times$8  & 128$\times$8$\times$8  & -                   \\ \cline{2-5}
		& dropout & 128$\times$8$\times$8  & 128$\times$8$\times$8  & 0.2                 \\ \hline
		\multirow{3}{*}{6} & conv    & 128$\times$8$\times$8  & 128$\times$8$\times$8  & 3$\times$3                 \\ \cline{2-5}
		& relu    & 128$\times$8$\times$8  & 128$\times$8$\times$8  & -                   \\ \cline{2-5}
		& pool    & 128$\times$8$\times$8  & 128$\times$4$\times$4  & 2$\times$2                 \\ \hline
		\multirow{2}{*}{7} & flatten & 128$\times$4$\times$4  & 2048     & -                   \\ \cline{2-5}
		& dropout & 2048     & 2048     & 0.2                 \\ \hline
		\multirow{2}{*}{8} & dense   & 2048     & 1024     & -                   \\ \cline{2-5}
		& relu    & 1024     & 1024     & -                   \\ \hline
		\multirow{2}{*}{9} & dense   & 1024     & 64       & -                   \\ \cline{2-5}
		& relu    & 64       & 64       & -                   \\ \hline
	\end{tabular}
	\caption{Network architectures of $\mu(\textbf{x})$ and $\Sigma(\textbf{x})$ on CIFAR10. Note that $\mu(\textbf{x})$ and $\Sigma(\textbf{x})$ share the first 8 layers.}
		\label{mu_cifar}
\end{table}

\begin{table}[h]
	\centering

	\begin{tabular}{l||l|l|l|l}
		\hline
		layer              & op.     & size-in  & size-out & kernel/dropout rate \\ \hline\hline
		\multirow{2}{*}{1} & dense    & 64  & 512 & -                \\ \cline{2-5}
		& dropout & 512 & 512 & 0.2                 \\ \hline
		\multirow{2}{*}{2} & dense    & 512 & 10 & -                 \\ \cline{2-5}
		& softmax    & 10 & 10 & -                   \\ \cline{2-5}
                \hline

	\end{tabular}
	\caption{Network architecture of $\Phi(\textbf{x})$ on CIFAR10.}
		\label{phi_cifar}
\end{table}

\begin{table}[h]
	\centering
	\begin{tabular}{l||l|l|l|l}
		\hline
		layer              & op.     & size-in  & size-out & kernel/dropout rate \\ \hline\hline
		\multirow{2}{*}{1} & conv    & 1$\times$28$\times$28  & 1$\times$28$\times$28 & 2$\times$2                 \\ \cline{2-5}
		& relu    & 1$\times$28$\times$28 & 1$\times$28$\times$28 & -                   \\ \cline{2-5} \hline
		\multirow{2}{*}{2} & conv    & 1$\times$28$\times$28 & 64$\times$14$\times$14 & 2$\times$2                 \\ \cline{2-5}
		& relu    & 64$\times$14$\times$14 & 64$\times$14$\times$14 & -                   \\ \cline{2-5} \hline
		\multirow{2}{*}{3} & conv    & 64$\times$14$\times$14 & 64$\times$14$\times$14 & 3$\times$3                 \\ \cline{2-5}
& relu    & 64$\times$14$\times$14 & 64$\times$14$\times$14 & -                   \\ \cline{2-5} \hline
		\multirow{2}{*}{4} & conv    & 64$\times$14$\times$14 & 64$\times$14$\times$14 & 3$\times$3                 \\ \cline{2-5}
& relu    & 64$\times$14$\times$14 & 64$\times$14$\times$14 & -                   \\ \cline{2-5} \hline
		\multirow{1}{*}{5} & flatten & 64$\times$14$\times$14  & 12544   & -                   \\ \cline{2-5}  \hline
		\multirow{2}{*}{6} & dense   & 12544     & 128     & -                   \\ \cline{2-5}
		& relu    & 128       & 128       & -                   \\ \hline		
		\multirow{2}{*}{7} & dense   & 128     & 64       & -                   \\ \cline{2-5}
		& relu    & 64       & 64       & -                   \\ \hline
	\end{tabular}
	\caption{Network architectures of $\mu(\textbf{x})$ and $\Sigma(\textbf{x})$ on MNIST. Note that $\mu(\textbf{x})$ and $\Sigma(\textbf{x})$ share the first 6 layers.}
		\label{mu_mnist}
\end{table}

\begin{table}[h]
	\centering

	\begin{tabular}{l||l|l|l|l}
		\hline
		layer              & op.     & size-in  & size-out & kernel/dropout rate \\ \hline\hline
		\multirow{2}{*}{1} & dense    & 64  & 128 & -                \\ \cline{2-5}
		& dropout & 128 & 128 & 0.5                 \\ \hline
		\multirow{2}{*}{2} & dense    & 128 & 10 & -                 \\ \cline{2-5}
		& softmax    & 10 & 10 & -                   \\ \cline{2-5}
		\hline

	\end{tabular}
	\caption{Network architecture of $\Phi(\textbf{x})$ on MNIST.}
		\label{phi_mnist}
\end{table}

\begin{table}[h]
	\centering

	\begin{tabular}{l||l|l|l|l}
		\hline
		layer              & op.     & size-in  & size-out & kernel/dropout rate \\ \hline\hline
		\multirow{1}{*}{1} & Alex-Net-fc7    & 3$\times$224$\times$224  & 4096 & -                 \\ \cline{2-5} \hline
		\multirow{2}{*}{2} & dense    & 4096 & 4096 & -                 \\ \cline{2-5}
& relu    & 4096 & 4096 & -                   \\ \cline{2-5} \hline
		\multirow{2}{*}{3} & dense    & 4096 & 4096 & -                 \\ \cline{2-5}
& relu    & 4096 & 4096 & -                   \\ \cline{2-5} \hline
		\multirow{2}{*}{4} & dense    & 4096 & 1024 & -                 \\ \cline{2-5}
& relu    & 1024 & 1024 & -                   \\ \cline{2-5} \hline
		\multirow{2}{*}{5} & dense    & 1024 & 512 & -                 \\ \cline{2-5}
& relu    & 512 & 512 & -                   \\ \cline{2-5} \hline

	\end{tabular}
	\caption{Network architectures of $\mu(\textbf{x})$ and $\Sigma(\textbf{x})$ on NUSWIDE LITE. Note that $\mu(\textbf{x})$ and $\Sigma(\textbf{x})$ share the first 4 layers.}
		\label{mu_nuswide}
\end{table}

\begin{table}[ht]
	\centering

	\begin{tabular}{l||l|l|l|l}
		\hline
		layer              & op.     & size-in  & size-out &    kernel/dropout rate             \\ \hline
		\multirow{2}{*}{1} & dense    & 512 & 81 & -                 \\ \cline{2-5}
		& softmax    & 81 & 81 & -                   \\ \cline{2-5}
		\hline

	\end{tabular}
	\caption{Network architecture of $\Phi(\textbf{x})$ on NUSWIDE LITE.}
		\label{phi_nuswide}
\end{table}

\begin{table}[ht]
	\centering

	\begin{tabular}{l||l|l|l|l}
		\hline
		layer              & op.     & size-in  & size-out & kernel/dropout rate \\ \hline\hline
		\multirow{1}{*}{1} & GooLeNet    & 3$\times$224$\times$224  & 1025 & -                 \\ \cline{2-5} \hline
		\multirow{2}{*}{2} & dense    & 1025 & 1024 & -                 \\ \cline{2-5}
		& relu    & 1024 & 1024 & -                   \\ \cline{2-5} \hline
		\multirow{2}{*}{3} & dense    & 1024 & 512 & -                 \\ \cline{2-5}
		& relu    & 512 & 512 & -                   \\ \cline{2-5} \hline
		\multirow{2}{*}{4} & dense    & 512 & 512 & -                 \\ \cline{2-5}
		& relu    & 512 & 512 & -                   \\ \cline{2-5} \hline

	\end{tabular}
	\caption{Network architectures of $\mu(\textbf{x})$ and $\Sigma(\textbf{x})$ on AWA. Note that $\mu(\textbf{x})$ and $\Sigma(\textbf{x})$ share the first 3 layers.}
		\label{mu_awa}
\end{table}

\begin{table}[ht]
	\centering

	\begin{tabular}{l||l|l|l|l}
		\hline
		layer              & op.     & size-in  & size-out &    kernel/dropout rate             \\ \hline
		\multirow{2}{*}{1} & dense    & 512 & 85 & -                 \\ \cline{2-5}
		& relu    & 85 & 85 & -                   \\ \cline{2-5}
		& normalize    & 85 & 85 & -                   \\ \cline{2-5}
		\hline

	\end{tabular}
	\caption{Network architecture of $\Phi(\textbf{x})$ on AWA.}
		\label{phi_awa}
\end{table}

\end{document}